\documentclass{INTERSPEECH2023}

\interspeechcameraready

\title{Mitigating Catastrophic Forgetting for Few-Shot Spoken Word Classification Through Meta-Learning}

\name{Ruan van der Merwe$^1$  and Herman Kamper$^2$}
\address{
  $^1$ByteFuse\\
  $^2$E\&E Engineering, Stellenbosch University, South Africa}
\email{ruan@bytefuse.ai, kamperh@sun.ac.za}

\usepackage{booktabs}
\usepackage{tabularx}
\usepackage{multirow}

\newcolumntype{C}{>{\centering\arraybackslash}X}
\newcolumntype{L}{>{\raggedright\arraybackslash}X}

\usepackage[prependcaption,textsize=scriptsize]{todonotes}
\definecolor{mycolor}{HTML}{FF6600}

\begin{document}

\maketitle
 
\begin{abstract}
We consider the problem of few-shot spoken word classification in a setting where a model is incrementally introduced to new word classes. This would occur in a user-defined keyword system where new words can be added as the system is used. In such a continual learning scenario, a model might start to misclassify earlier words as newer classes are added, i.e.\ catastrophic forgetting. To address this, we propose an extension to model-agnostic meta-learning (MAML). In our new approach, each inner learning loop---where a model ``learns how to learn'' new classes---ends with a single gradient update using stored templates from all the classes that the model has already seen (one template per class). We compare this method to OML (another extension of MAML) in few-shot isolated-word classification experiments on Google Commands and FACC. Our method consistently outperforms OML in experiments where the number of shots and the final number of classes are varied.
\end{abstract}

\section{Introduction}

Imagine a speech system that a user can teach new commands by providing it with just a few examples per word class. To start out with, the user might provide the system with examples of the words ``sing'', ``open'' and ``close'', and with just a handful of support examples, the system should be able to correctly classify new test inputs. (This should work irrespective of the language of the user.) In contrast to conventional speech recognition systems that are trained on thousands of hours of examples, such a system would be \textit{few-shot}. Inspired by the observation that humans can learn new words from very few examples, a number of studies in machine learning have started to look at this problem of few-shot word classification~\cite{lake2014one,elof_herman_engelbrecht_icassp,kao2023efficiency}.

But now imagine that, as the user is using the system, they want to add more words to the system, e.g. ``turn'' and ``give''. As more and more words are added, the system might start to misclassify words that it learned earlier---the problem of catastrophic forgetting~\cite{mccloskey1989catastrophic,goodfellow2013empirical}. The combination of dynamic environments, limited support examples used for training, and continual learning make this task a major challenge. While other studies have look at the few-shot problem~\cite{lake2014one,heggan2022metaaudio}, the proposed methods do not deal with the continual learning problem. In this paper we propose a new approach for few-shot continual learning and evaluate it specifically for isolated word classification.

Outside of speech processing, there has been several studies on continual learning, e.g.~\cite{mai2022online}. Many of these studies try to explicitly address the problem of catastrophic forgetting~\cite{li2017learning, lopez2017gradient}. Within speech research, there has been some limited attempts to address the continual learning problem, specifically in automatic speech recognition (ASR)~\cite{sadhu2020continual} and keyword spotting applications~\cite{huang2022progressive}. However, these studies do not consider the few-shot learning setting, but rather on adding new vocabulary words to supervised models trained on substantial amounts of labelled data. Within the signal processing community, there has been some studies looking at both few-shot learning and continual updating~\cite{wang2021few}, but this was for general audio and not spoken word classification.

In this paper we specifically look at addressing few-shot continual learning by utilising meta-learning techniques, where algorithms learn automatically how to solve the continual learning task~\cite{javed2019meta, gupta2020look, beaulieu2020learning}.
We specifically extend model-agnostic meta-learning (MAML)~\cite{finn2017model}, which is a meta-learning technique that optimises an initial set of model weights such that they can be quickly updated to a new task.
MAML has been used before within speech research for speaker adaptive training~\cite{klejch2019speaker}, and data-efficient ASR \cite{indurthi2019data, winata20_interspeech}, but not for few-shot continual word learning.

We propose a new approach: MAML for \textbf{con}tinual learning (MAMLCon).
This extension over MAML is very simple, but it leads to consistent improvements in few-shot word classification.
MAMLCon specifically extends MAML by explicitly doing meta-learning of an increasing number of new classes in the inner loop of the algorithm. At the end of the inner loop, MAMLCon also performs a single update using templates stored for all the classes seen up to that point. Since MAMLCon has learned how to learn continually, it is able to do so efficiently at test time on classes that are completely unseen during meta-learning.

We compare MAMLCon to another continual learning extension of MAML called OML~\cite{javed2019meta}. We perform experiments where we vary the number of shots, the number of steps where classes are added, and the final number of word classes. In all cases the simple MAMLCon extension outperforms OML in isolated word %continual
few-shot classification.
 
\section{MAMLCon}

\subsection{Background on MAML} \label{sec:maml-background}

Model-agnostic meta-learning (MAML)~\cite{finn2017model} is an algorithm that aims to learn an initial set of weights that can be rapidly adapted to new tasks using just a few examples
from the target task.
Consider
the example of one-shot speech classification.
We want a model that can learn to classify new words based on a single training example per word class. E.g.\ we give the model a \textit{support set}\footnote{In few-shot classification, the \textit{support set} is the small set of training examples that we get for the target task.} with a single example for ``sing'', ``open'', ``close'' and then want the model to accurately classify test inputs from one of these classes.
A naive approach would be to start with a randomly initialised model and then simply update its weights through gradient descent directly on this support set.
The idea behind MAML is to instead learn good initial weights which can then subsequently be fine-tuned.
MAML does this by using a large labelled dataset and then simulating many few-shot classification tasks.
Continuing with our example, let's say we have a very large training set of isolated words with their labels (no examples from our few-shot classes). From this training dataset we can sample a meta-support set and a meta-test set, e.g. ``hello'', ``drop'', ``greetings''. In the so-called \textit{inner loop} of the MAML algorithm, we then update the model weights using a few gradient descent steps on the support set.
Instead of storing the resulting weights from these inner-loop updates, MAML optimises the initial weights $\theta$ on top of which the inner-loop updates are performed.
I.e., the \textit{outer loop} of MAML tries to find a good initialisation for doing a few gradient steps on a handful of examples.
The result is weights $\theta^*$ that are optimised so that they work well when a few gradient steps are applied on top of them using a small set of support examples.

More formally, in the inner loop, the model's current weights at step $j$, $\theta^j_{0}$, are optimised for a given task $\mathcal{T}_i$, resulting in updated weights $\theta^j_{T}$, where $T$ is the total number of inner-loop update steps. In the outer loop, the performance of the fine-tuned model %,
$\theta^j_{T}$
is evaluated
% ,
on a meta-test set,
and the initial weights $\theta^j_{0}$ are then updated through:
\begin{equation}
\label{eq:maml-opt-form}
\theta^{j+1}_{0} \leftarrow \theta^j_{0} - \beta \nabla_{\theta^{j}_{0}} \sum_{\mathcal{T}_i} \mathcal{L}_{\mathcal{T}_i}\left(X^{\textrm{TEST}}_{i}, Y^{\textrm{TEST}}_{i}, \theta^j_{T}\right)
\end{equation}
Here, $X_i$ and $Y_i$ are data points from task $\mathcal{T}_i$, and $\beta$ is the outer learning rate, with the inner-loop update steps having an inner learning rate $\alpha$. Updating $\theta^j_{0}$ in this manner leads to optimised weights $\theta^*$ which can be fine-tuned to new tasks in only a few steps. When the inner loop is constrained to only a few examples per class, the algorithm can learn to accomplish the task with a limited number of examples, thus resulting in a few-shot classification model.

\begin{figure}[!t]
  \centering
  \includegraphics[width=\linewidth]{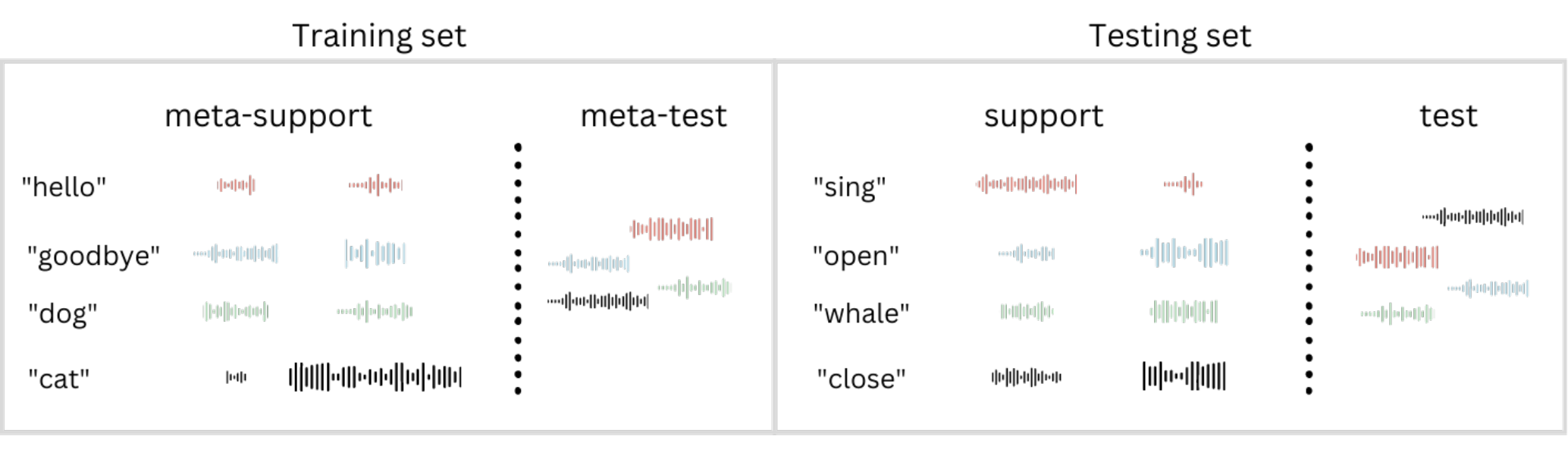}
  \caption{During training, MAML samples meta-support and \mbox{-test} sets from labelled data. At test time, it is then presented with a support set containing classes never seen during training, and asked to classify test items from these classes.}
  \label{fig:example-dataset}
\end{figure}

To test a model after training it using MAML, we
% After successfully training the model using the MAML algorithm, one
can sample multiple groups of words from our few-shot classes and construct multiple scenarios  %scenarious
where you train on a support set and measure on a held-out test set. The optimised model %,
$\theta^*$ is copied to each distinct scenario %scenarios
for training. An example of how these meta-training and -testing scenarios are constructed is shown in Figure~\ref{fig:example-dataset}, where we 
show just
% sample
one task in both the training and testing stages. 
For further reading on meta-learning and MAML, please refer to \cite{tian2022meta}.

\subsection{MAMLCon: Learning to Continually Learn}

\begin{figure}[th]
\centering
  \includegraphics[width=\linewidth]{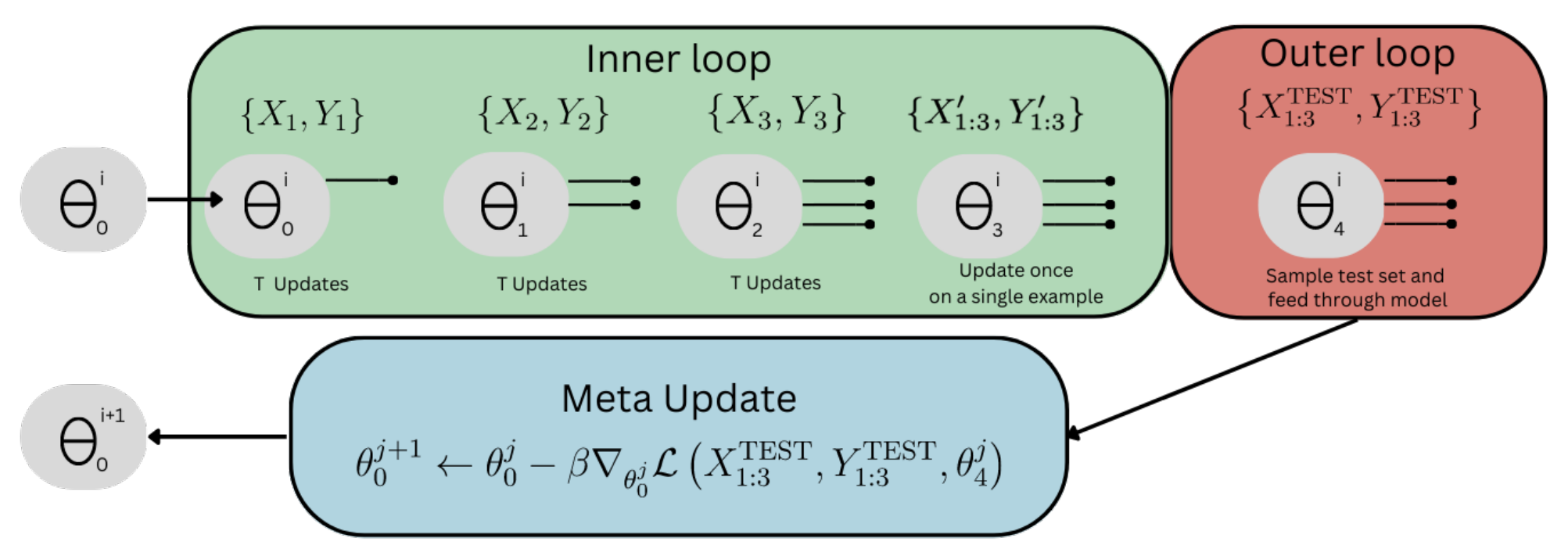}
    \caption{The MAMLCon training process. We construct the continual learning  setup directly as a meta-task, where the algorithm is tasked to learn how to perform well in continual learning setup while being allowed to observe one already seen example from previously learned word groups and update its weights with one update step.}
  \label{fig:mamlcon_diagram}
\end{figure}

Consider the following example for word classification in a continual learning setting.
Let's say at test time a model has received a support set for the words
``sing'', ``open'' and ``close''.
We used MAML and updated the model on this support set and it achieves reasonable performance.
But now we want the model to additionally be able to classify the words ``turn'' and ``give''.
We give the model a few more support examples for these new words and update its weights through further fine-tuning.
Later on, we want to add even more words by just giving a few examples.
The problem is that as we add more and more words, the model would start to fail on words that it learned earlier. This is called catastrophic forgetting.

To address this, we propose a new extension of MAML:
\textbf{m}odel-\textbf{a}gnostic \textbf{m}eta-\textbf{l}earning for \textbf{con}tinual learning~(MAMLCon).
MAMLCon extends MAML in two ways. First, it formulates the continual learning problem itself
as a meta-learning task.
Secondly, it utilises %utilise
a single update step on previously acquired knowledge.
The motivation for this step is to optimise the model such that one can use use the smallest possible dataset (one example per class) to maintain performance on previously learned words.

The training process of MAMLCon is shown in Figure \ref{fig:mamlcon_diagram}.
As an example,
let's say that during training we sample a meta-support set consisting of five examples each for ``hello'', ``drop'', ``greetings''. 
In MAML we would just fine-tune on all the examples together. Instead, in
the inner-loop training phase of MAMLCon, the model is first trained for $T$ steps on the ``hello'' examples, followed by $T$ steps of training on ``drop'' and then $T$ steps %training
on ``greetings''.
Once the model has been trained on all examples in the meta-support set, a single batched weight update step is performed
using a single stored example for each of the ``hello'', ``drop'', ``greetings'' classes.
In the outer loop, the meta-test set, which contains samples for all words in the meta-support set, is used to evaluate the performance of the model, % on all encountered words,
and the original weights are updated %based on this performance
to obtain an optimal set of weights %, denoted as 
$\theta^*$.
Because with MAMLCon the model has seen incremental learning during training, these
% These
weights are optimised to facilitate few-shot continual learning. 
This means we can the update the model further on ``turn'' and ``give'' and the model would still perform well on ``hello'', ``drop'' and ``greetings''.

To state this formally, in the inner loop, the model's weights, $\theta_0^j$, are updated through sequential training on new classes in the meta-support set. The inner-loop optimisation is performed through the calculation of gradients with respect to $\theta_{i}^j$ based on the loss computed on a per-class (or per-group of classes) basis from the meta-support set, leading to the updated weights $\theta_{i+1}^{j}$.
At the end of the inner loop, a single %After all classes have been trained, the model is updated with a single
weight update is performed on a previously seen template from each class, enabling the model to leverage its prior knowledge. 
In Figure~\ref{fig:mamlcon_diagram}, this set of templates is denoted with a dash, $\{ X_{1:3}', Y_{1:3}'\}$. 
The outer loop computes the loss on the meta-test set and applies the meta-update step to %update
the original weights to obtain $\theta_{0}^{j+1}$. The update is performed based on the gradient of the test loss with respect to $\theta_0^j$, as %expressed
in Equation \ref{eq:maml-opt-form}. 

At test time, MAMLCon is used by just following the inner loop. 
% This is depicted in Figure \ref{fig:mamlcon_rw}, where $\theta^*_0$ represents the weights of the model trained with MAMLCon.
Every time that classes are added, $T$ update steps are followed with one update step on a set of templates for all classes learned up to that point.
This means that in a real-life use case, we will just have to store a single example per class to act as templates in future updates.

Our method is most similar to online aware meta-learning (OML)~\cite{javed2019meta}. 
The OML classifier consists of
a feature extractor with weights $\theta_{\text{FE}}$ %,
that feeds % feed
into a prediction network with weights %(
$\theta_{\text{PN}}$. %).
% During
In OML's the inner loop, they %then
sample $N$ classes to train on sequentially but only update %the 
$\theta_{\text{PN}}$, leading to $\theta^*_{\text{PN}}$. After training these $N$ classes, they sample a random batch of data and measure the meta-test loss on this batch. They then backpropogate through this entire process to update $\theta_{\text{FE}}$ and $\theta_{\text{PN}}$. 
Our %proposed
method differs from OML in several ways. Firstly,
in the inner loop,
% during the inner-loop training phase,
we update the entire network and not just the prediction network. % Additionally,
Secondly,
we allow the model to access a %one
single example of a previously seen class during the inner-loop training phase. Finally, in %In
contrast to OML, % the existing method,
we do not perform the meta-test on a random sample of classes, but instead on all classes seen up to that point. 

\section{Experimental Setup}

\textbf{Data.} We perform word classification experiments using the Flickr 8k Audio Caption Corpus (FACC)~\cite{harwath2015deep} and the Google Commands v2 dataset~\cite{warden2018speech}.
For the experiments on FACC, utterances
are
% were
segmented into isolated words using forced alignments, and words with the same stem are %were
grouped into a single class. 
Both the FACC and Google Commands datasets are %were
split so that words with the same stem will not appear in the training and test %evaluation
sets. 
For FACC, this results in approximately 100 unique stems that can be sampled %to create unique groups of words
for continual learning, while there are 10 unique stems for Google Commands. 
We divide these stems randomly into our test and train splits.
Between epochs in meta-learning, the same word class will be assigned a different integer label so that the model is not able memorise a particular word in the meta-learned weights. 

\textbf{Models.}
All words are parameterised as mel-frequency cepstral coefficients (MFCCs) with delta and delta-delta features.
% To ensure consistency in length, the i
Input items are zero-padded to a consistent length. A simple 3-layer 2D convolutional neural network is applied to extract features from the MFCCs, which are then fed into a single fully connected layer that is trained to classify the given words. We use the same architecture for OML. % to ensure a fair comparison. 
The Adam optimiser \cite{KingmaB14} is used for both inner and outer loop updates, %adaptation steps,
with a learning rate of $0.001$ for the inner loop % weight updated,
and $0.0001$ for the outer loop. % outer-loop weight updates.

In all the experiments below we start with a set of initial words, and then incrementally add more word classes.
For the initial set of words being learned by the model, we 
perform %performed
$T=30$ weight updates to ensure saturation of the model to simulate the scenario in the real world of having a well-trained model and subsequently updating it. After this, for each new group of classes added to the model, $T=5$ update steps are performed. In the quick adaptation step on the templates at the end of the inner loop, a single example per class is sampled from the %meta-
support set and a single update is performed.
% ; whereafter only one update step, $T=1$, is performed.
We use %also implement
the first-order MAML algorithm \cite{finn2017model}, which ignores the meta-learning process's second-order derivatives; this doesn't affect performance
while speeding up computation and reducing memory requirements \cite{finn2017model, nichol2018first}.
We adapt the Learn2Learn software package \cite{Arnold2020-ss} for training both OML and MAMLCon.\footnote{Source code: \url{https://github.com/ByteFuse/MAMLCon}}

\textbf{Evaluation.}
We consider different continual learning scenarios.
All start with an initial set of few-shot learned word classes: this number of initial classes are denoted as \textbf{CS}.
We then incrementally introduce a number of additional word types (\textbf{CA}) at every update step.
The final number of word types is denoted as $N$.
An experiment can then be summarised using a succinct notation: e.g.\ $N$50:\textbf{CS}5:\textbf{CA}5 would represent a
a scenario in which the model ends with
% is trained on
a total of 50 word classes, with each iteration incorporating five new words after initially training on five words.

\section{Experiments}

We compare MAMLCon to OML for few-shot word classification in a range of continual learning experiments. We do not evaluate MAML in isolation, as it has been surpassed in performance by OML and other recent advancements \cite{javed2019meta, gupta2020look}.

\subsection{Frequent vs Infrequent Updates}

A good continual learning algorithm should perform well in scenarios where
we add many words at every update step (therefore requiring fewer updates to reach the final number of types $N$) as well as scenarios where a small number of words are added at every update (requiring more frequent updates to reach $N$). 
We compare MAMLCon to OML in both these scenarios, referred to, respectively, as infrequent and frequent updates.
For infrequent updates
we consider these setups:
% , the following setups are 
% used:
$N5$:\textbf{CS}1:\textbf{CA}3, $N10$:\textbf{CS}2:\textbf{CA}5 and $N50$:\textbf{CS}5:\textbf{CA}20. For %the
frequent updates we consider %, the following setups were used:
$N5$:\textbf{CS}1:\textbf{CA}1, $N10$:\textbf{CS}2:\textbf{CA}1 and $N50$:\textbf{CS}5:\textbf{CA}5. All setups here use $K=5$ shots (we vary this in the section below).

\begin{table}[!b]
\centering
\small
\caption{Few-shot classification accuracy (\%) over all $N$ classes for continual learning settings where a small number of classes are added frequently, or a large number of classes are added infrequently. $N$ is the final number of classes after continual learning.}
\label{tab:overall_results}
    \renewcommand{\arraystretch}{1.1}
    \footnotesize
    \begin{tabularx}{\linewidth}{@{}lCCCC@{}}
        \toprule
        & \multicolumn{2}{c}{Google Commands} & \multicolumn{2}{c}{FACC} \\
        \cmidrule(lr){2-3} \cmidrule(l){4-5}
        $N$ final classes & 5 & 10 & 10 & 50 \\
        \midrule
        \underline{\textit{Infrequent updates:}} \\
        OML & 62.1 & 49.9 & 72.8 & 32.3 \\
        MAMLCon & \textbf{85.2} & \textbf{73.6} & \textbf{86.7} & \textbf{74.5} \\
        \addlinespace
        \underline{\textit{Frequent updates:}} \\
        OML & 61.8 & 36.5 & 75.1 & 51.8 \\
        MAMLCon & \textbf{82.7} & \textbf{72.9}  & \textbf{76.8} & \textbf{71.7} \\
        \bottomrule
    \end{tabularx}
\end{table}

The results are shown %can be found
in Table~\ref{tab:overall_results}, where $N$ is used to identify the particular learning scenario.
By looking at the infrequent update scenario,
% Upon analysing the infrequent update scenario,
we observe that MAMLCon 
%can
achieves high %levels of
accuracies
% accuracy 
in both smaller ($N=\{5,10\}$) and larger class scenarios ($N=50$).
In contrast, OML struggles particularly when more classes need to be learned: this
can be seen  when looking at the sharp drop in accuracy between the results for the FACC dataset in the $N=10$ and $N=50$ cases, %case and
and
on the Google Commands dataset when going from $N=5$ to %and 
$N=10$. 
A similar pattern emerges in the frequent update scenario, where we see that OML shows
% relatively
large drops in accuracy
when learning more classes: a particularly large drop is observed on Google Commands when going from $N = 5$ to $N = 10$.
Overall, the results demonstrate the superior performance of MAMLCon over OML in both frequent and infrequent update scenarios.

\subsection{Few-shot Capabilities}

\begin{figure}[!t]
  \centering
  \includegraphics[width=\linewidth]{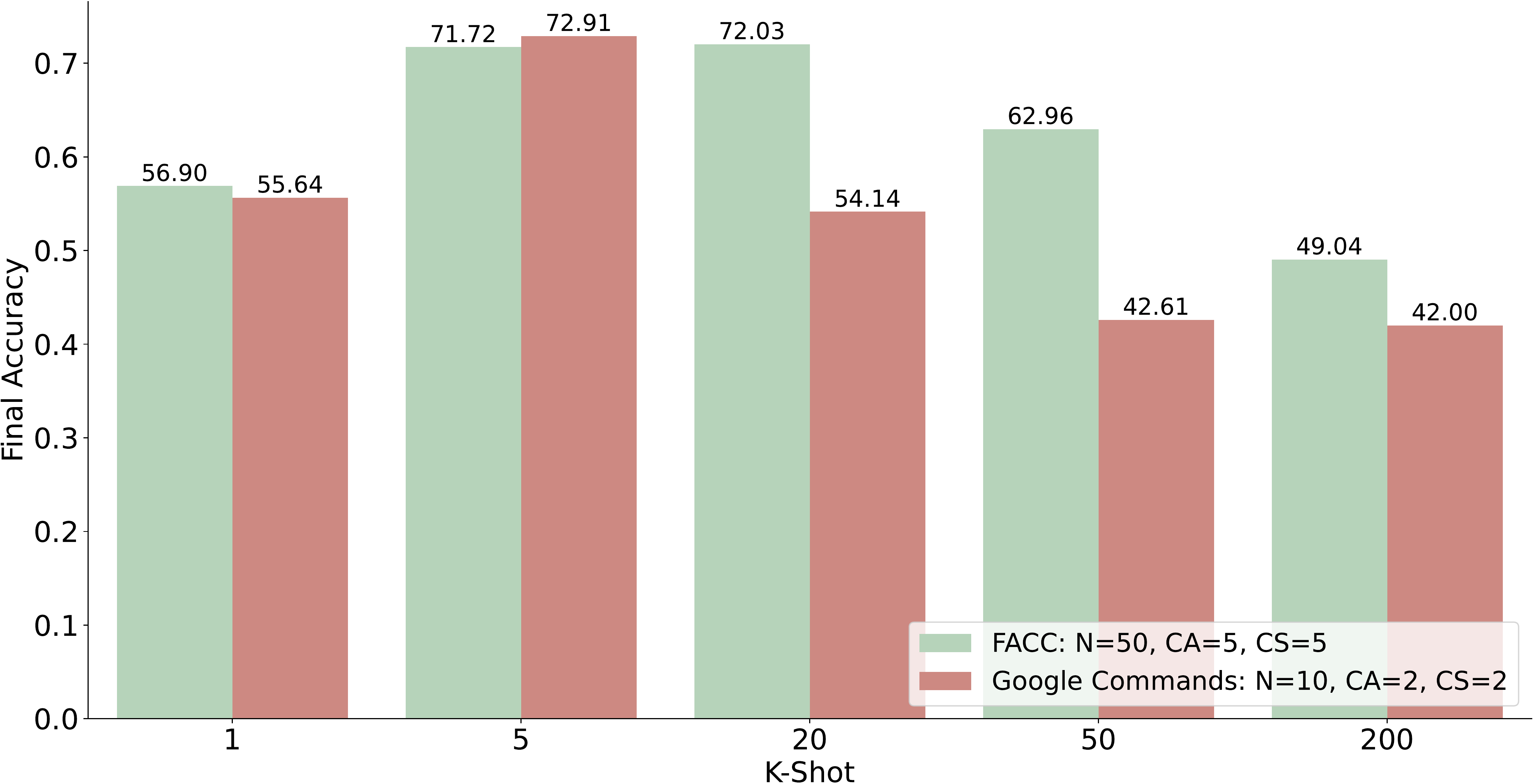}
  \caption{Few-shot classification accuracy (\%) of MAMLCon as the number of shots $K$ per class is varied.}
  \label{fig:k_shot_investigation}
\end{figure}

The number of
support
% training
examples a model can use %requires
for learning a new word 
would depend on the specific practical setting: in some cases we would have only one example per class, while in other cases we could get substantially more.
Here
we assess the performance of MAMLCon 
as the number $K$ of support examples (the number of ``shots'') are varied.
We investigate how well MAMLCon operates under these different conditions to gain a better understanding of its capabilities. 

Concretely,
% Below,
we present the performance for continual learning setups of $N50$:\textbf{CS}5:\textbf{CA}5 when evaluating on the FACC dataset and $N10$:\textbf{CS}2:\textbf{CA}1 when evaluating on the Google Commands 
dataset %,
over different values of $K$. These setups were chosen as they represent the most challenging scenarios, requiring multiple weight updates between the initial and final classes.

As seen in Figure \ref{fig:k_shot_investigation}, when focusing solely on the results for the FACC dataset, %it can be observed that
as $K$ increases from $1$ to $20$, the overall performance improves as expected, with only a small increase in performance between $K=5$ and $K=20$. 
However, as $K$ continues to increase, performance decreases. This pattern is also evident 
in the Google Commands %dataset
results. 

It is encouraging that MAMLCon still performs well with a small number of shots, but it is also somewhat surprising and concerning that as $K$ increases, performance starts to deteriorate. This
% The observation of the
relationship between accuracy and the number of training examples in Figure \ref{fig:k_shot_investigation} can be explained by the trade-off between sample complexity and catastrophic forgetting. We speculate that a moderate value of $K$, in the range of 20, is sufficient to acquire a robust representation of the task at hand, which is to learn a new word. However, as $K$ increases beyond this point, the weight updates for the new classes may become excessive, resulting in the model forgetting previously learned information. 

\subsection{Retention of Knowledge}

In the preceding sections 
we looked at performance across all words after a few-shot system has been trained in a continual learning setting.
But how does performance differ between words that are learned earlier relative to words added later in the continual learning cycle? To answer this, we look at the performance of individual words.
This allows us to determine how well the model performs on previous classes and how well it retains the knowledge about those words after being trained on new words.

Table \ref{tab:retention_results} shows the results of MAMLCon, OML %,
and a model which was not pre-trained%,
on the FACC dataset.
We use a $N50$:\textbf{CS}5:\textbf{CA}5 setup, with $K=20$. 
This means that there will be ten update steps, with five word classes being added each time.
The performance for the words learned in the very first group are given in the row with the 1-5 label, while the words learned in the very last update are given in the 46-50 row.
The accuracy after initial training (S) and the final training (E) for each label group is displayed, along with the difference ($\Delta$) between these two accuracy scores.

MAMLCon again outperforms OML in terms of overall accuracy, achieving 77.0\% accuracy versus the 64.5\% of OML. 
Looking at individual words,
MAMLCon is effective in retaining its knowledge of early label groups (1-30) while struggling more to maintain its accuracy over the later label groups. Conversely, OML performs better in retaining knowledge over later label groups, but shows low accuracy for the early groups.

\begin{table}[!t]
\centering
\small
\caption{Evaluation of knowledge retention capabilities in continual learning models on the FACC dataset. We measure the accuracy for each label group as it was trained, as well as at the end of training after all words have been learned by the model. We then show the the difference ($\Delta$) between the start (S) and end (E) accuracies. The final accuracy when taking all labels into account is also shown.}
\label{tab:retention_results}
    \renewcommand{\arraystretch}{1.1}
    \footnotesize
    \begin{tabularx}{\linewidth}{@{}lCCCCCC@{}}
        \toprule
        & \multicolumn{2}{c}{MAMLCon} & \multicolumn{2}{c}{OML} & \multicolumn{2}{c}{No Pre-Training} \\
        \cmidrule(lr){2-3} \cmidrule(lr){4-5} \cmidrule(l){6-7}
        Labels & S/E & $\Delta$ & S/E & $\Delta$ & S/E & $\Delta$ \\
        \midrule
        1-5 & 95/95   & 0  & 100/35  & -65 & 90/20  & -70 \\
        6-10 & 100/95 & -5  & 85/5   & -80 & 85/50  & -35 \\
        11-15 & 90/85 & -5  & 100/70 & -30 & 90/25  & -65 \\
        16-20 & 95/70 & -25 & 90/75  & -25 & 100/40 & -60 \\
        21-25 & 95/80 & -15 & 75/65  & -10 & 60/10  & -50 \\
        26-30 & 95/70 & -25 & 100/95 & -5  & 100/40 & -60 \\
        31-35 & 95/50 & -45 & 80/55  & -25 & 85/20  & -65 \\
        36-40 & 80/70 & -10 & 75/80  & 5 & 90/0     & -90 \\
        41-45 & 85/60 & -25 & 90/75  & -15 & 75/60  & -15 \\
        46-50 & -/95  & -   & -/90   & -  & -/65   & - \\
        \midrule
        Accuracy & 77.0 & \multicolumn{1}{l}{} & 64.5 &  & 33.0 & \multicolumn{1}{l}{} \\ 
        \bottomrule
    \end{tabularx}
\end{table}

\section{Conclusion}

We
proposed 
a novel few-shot continual learning algorithm: model-agnostic meta-learning for continual learning (MAMLCon).
It is an extension of MAML that
formulates the few-shot continual learning  %continual learning
task as a meta-task, allowing the weights to be updated only once by a previously seen word example upon completion of training on new words.
We compared MAMLCon to OML, a previous meta-learning algorithm for continual learning.
The findings show that MAMLCon outperforms OML in %regarding
overall accuracy
across two %various
datasets and label distribution sizes under both infrequent and frequent update scenarios. Furthermore, our results indicate that MAMLCon effectively maintains knowledge of early label groups while showing more difficulty retaining knowledge of later groups. Nonetheless, it achieves a higher overall accuracy. 

\newpage
\bibliographystyle{IEEEtran}
\bibliography{mybib}

\begin{thebibliography}{10}
\providecommand{\url}[1]{#1}
\csname url@samestyle\endcsname
\providecommand{\newblock}{\relax}
\providecommand{\bibinfo}[2]{#2}
\providecommand{\BIBentrySTDinterwordspacing}{\spaceskip=0pt\relax}
\providecommand{\BIBentryALTinterwordstretchfactor}{4}
\providecommand{\BIBentryALTinterwordspacing}{\spaceskip=\fontdimen2\font plus
\BIBentryALTinterwordstretchfactor\fontdimen3\font minus
  \fontdimen4\font\relax}
\providecommand{\BIBforeignlanguage}[2]{{%
\expandafter\ifx\csname l@#1\endcsname\relax
\typeout{** WARNING: IEEEtran.bst: No hyphenation pattern has been}%
\typeout{** loaded for the language `#1'. Using the pattern for}%
\typeout{** the default language instead.}%
\else
\language=\csname l@#1\endcsname
\fi
#2}}
\providecommand{\BIBdecl}{\relax}
\BIBdecl


\bibitem{lake2014one}
B.~Lake, C.-y. Lee, J.~Glass, and J.~Tenenbaum, ``One-shot learning of
  generative speech concepts,'' in \emph{Proceedings of the Annual Meeting of
  the Cognitive Science Society}, vol.~36, 2014.

\bibitem{elof_herman_engelbrecht_icassp}
R.~Eloff, H.~A. Engelbrecht, and H.~Kamper, ``Multimodal one-shot learning of
  speech and images,'' in \emph{International Conference on Acoustics, Speech
  and Signal Processing}, 2019.

\bibitem{kao2023efficiency}
W.-T. Kao, Y.-K. Wu, C.-P. Chen, Z.-S. Chen, Y.-P. Tsai, and H.-Y. Lee, ``On
  the efficiency of integrating self-supervised learning and meta-learning for
  user-defined few-shot keyword spotting,'' in \emph{Spoken Language Technology
  Workshop}, 2023.

\bibitem{mccloskey1989catastrophic}
M.~McCloskey and N.~J. Cohen, ``Catastrophic interference in connectionist
  networks: The sequential learning problem,'' in \emph{Psychology of learning
  and motivation}, 1989, vol.~24.

\bibitem{goodfellow2013empirical}
I.~J. Goodfellow, M.~Mirza, D.~Xiao, A.~Courville, and Y.~Bengio, ``An
  empirical investigation of catastrophic forgetting in gradient-based neural
  networks,'' \emph{arXiv preprint arXiv:1312.6211}, 2013.

\bibitem{heggan2022metaaudio}
C.~Heggan, S.~Budgett, T.~Hospedales, and M.~Yaghoobi, ``Metaaudio: A few-shot
  audio classification benchmark,'' in \emph{International Conference on
  Artificial Neural Networks}, vol. 13529, 2022.

\bibitem{mai2022online}
Z.~Mai, R.~Li, J.~Jeong, D.~Quispe, H.~Kim, and S.~Sanner, ``Online continual
  learning in image classification: An empirical survey,''
  \emph{Neurocomputing}, vol. 469, 2022.

\bibitem{li2017learning}
Z.~Li and D.~Hoiem, ``Learning without forgetting,'' \emph{Transactions on
  Pattern Analysis and Machine Intelligence}, vol.~40, 2017.

\bibitem{lopez2017gradient}
D.~Lopez-Paz and M.~Ranzato, ``Gradient episodic memory for continual
  learning,'' \emph{Advances in Neural Information Processing Systems},
  vol.~30, 2017.

\bibitem{sadhu2020continual}
S.~Sadhu and H.~Hermansky, ``Continual learning in automatic speech
  recognition,'' in \emph{Interspeech}, 2020.

\bibitem{huang2022progressive}
Y.~Huang, N.~Hou, and N.~F. Chen, ``Progressive continual learning for spoken
  keyword spotting,'' in \emph{International Conference on Acoustics, Speech
  and Signal Processing}, 2022.

\bibitem{wang2021few}
Y.~Wang, N.~J. Bryan, M.~Cartwright, J.~P. Bello, and J.~Salamon, ``Few-shot
  continual learning for audio classification,'' in \emph{International
  Conference on Acoustics, Speech and Signal Processing}, 2021.

\bibitem{javed2019meta}
K.~Javed and M.~White, ``Meta-learning representations for continual
  learning,'' \emph{Advances in Neural Information Processing Systems},
  vol.~32, 2019.

\bibitem{gupta2020look}
G.~Gupta, K.~Yadav, and L.~Paull, ``Look-ahead meta learning for continual
  learning,'' \emph{Advances in Neural Information Processing Systems},
  vol.~33, 2020.

\bibitem{beaulieu2020learning}
S.~Beaulieu, L.~Frati, T.~Miconi, J.~Lehman, K.~O. Stanley, J.~Clune, and
  N.~Cheney, ``Learning to continually learn,'' in \emph{European Conference on
  Artificial Intelligence}, 2020, vol. 325.

\bibitem{finn2017model}
C.~Finn, P.~Abbeel, and S.~Levine, ``Model-agnostic meta-learning for fast
  adaptation of deep networks,'' in \emph{International Conference on Machine
  Learning}, 2017.

\bibitem{klejch2019speaker}
O.~Klejch, J.~Fainberg, P.~Bell, and S.~Renals, ``Speaker adaptive training
  using model agnostic meta-learning,'' in \emph{Automatic Speech Recognition
  and Understanding Workshop}, 2019.

\bibitem{indurthi2019data}
S.~Indurthi, H.~Han, N.~K. Lakumarapu, B.~Lee, I.~Chung, S.~Kim, and C.~Kim,
  ``Data efficient direct speech-to-text translation with modality agnostic
  meta-learning,'' \emph{arXiv preprint arXiv:1911.04283}, 2019.

\bibitem{winata20_interspeech}
G.~I. Winata, S.~Cahyawijaya, Z.~Liu, Z.~Lin, A.~Madotto, P.~Xu, and P.~Fung,
  ``Learning fast adaptation on cross-accented speech recognition,'' in
  \emph{Interspeech}, 2020.

\bibitem{tian2022meta}
Y.~Tian, X.~Zhao, and W.~Huang, ``Meta-learning approaches for
  learning-to-learn in deep learning: A survey,'' \emph{Neurocomputing}, vol.
  494, 2022.

\bibitem{harwath2015deep}
D.~Harwath and J.~Glass, ``Deep multimodal semantic embeddings for speech and
  images,'' in \emph{Workshop on Automatic Speech Recognition and
  Understanding}, 2015.

\bibitem{warden2018speech}
P.~Warden, ``Speech commands: A dataset for limited-vocabulary speech
  recognition,'' \emph{arXiv preprint arXiv:1804.03209}, 2018.

\bibitem{KingmaB14}
D.~P. Kingma and J.~Ba, ``Adam: {A} method for stochastic optimization,'' in
  \emph{International Conference on Learning Representations}, 2015.

\bibitem{nichol2018first}
A.~Nichol, J.~Achiam, and J.~Schulman, ``On first-order meta-learning
  algorithms,'' \emph{arXiv preprint arXiv:1803.02999}, 2018.

\bibitem{Arnold2020-ss}
S.~M. Arnold, P.~Mahajan, D.~Datta, I.~Bunner, and K.~S. Zarkias,
  ``learn2learn: A library for meta-learning research,'' \emph{arXiv preprint
  arXiv:2008.12284}, 2020.

\end{thebibliography}

\end{document}